\documentclass{article}

\usepackage{microtype}
\usepackage{graphicx}
\usepackage{subfigure}
\usepackage{booktabs} % for professional tables

\usepackage[breaklinks]{hyperref}

\usepackage[accepted]{icml2023}

\usepackage{amsmath}
\usepackage{amssymb}
\usepackage{mathtools}
\usepackage{amsthm}
\usepackage{graphicx}
\usepackage{booktabs}
\usepackage[capitalize,noabbrev]{cleveref}

\theoremstyle{plain}

\theoremstyle{definition}

\theoremstyle{remark}

\usepackage{listings}
\usepackage{pythonhighlight}
\usepackage{hyperref}
\usepackage[T1]{fontenc}    % use 8-bit T1 fonts
\usepackage{url}            % simple URL typesetting
\usepackage{booktabs}       % professional-quality tables
\usepackage{amsfonts}       % blackboard math symbols
\usepackage{nicefrac}       % compact symbols for 1/2, etc.
\usepackage{microtype}      % microtypography
\usepackage{bbm}
\usepackage{amsmath}
\usepackage{enumerate}
\usepackage{empheq}
\usepackage{setspace}
\usepackage{amsmath,amssymb,amsthm,times,ifthen}
\usepackage{graphicx}
\usepackage{epstopdf}
\usepackage{epsf,epic,psfrag,mathabx}
\usepackage{enumerate}
\newlength\mytemplen
\newsavebox\mytempbox
\usepackage{wrapfig}
\makeatletter
\newcommand\mybluebox{%
    \@ifnextchar[%]
       {\@mybluebox}%
       {\@mybluebox[0pt]}}

\newcommand\mygreybox{%
    \@ifnextchar[%]
       {\@mygreybox}%
       {\@mygreybox[0pt]}}
       
\def\@mybluebox[#1]{%
    \@ifnextchar[%]
       {\@@mybluebox[#1]}%
       {\@@mybluebox[#1][0pt]}}

\def\@mygreybox[#1]{%
    \@ifnextchar[%]
       {\@@mygreybox[#1]}%
       {\@@mygreybox[#1][0pt]}}

\def\@@mybluebox[#1][#2]#3{
    \sbox\mytempbox{#3}%
    \advance\mytemplen #1\relax
    \ht\mytempbox\mytemplen
    \mytemplen\dp\mytempbox
    \advance\mytemplen #2\relax
    \dp\mytempbox\mytemplen
    \colorbox{myblue}{\hspace{1em}\usebox{\mytempbox}\hspace{1em}}}

\def\@@mygreybox[#1][#2]#3{
    \sbox\mytempbox{#3}%
    \mytemplen\ht\mytempbox
    \advance\mytemplen #1\relax
    \ht\mytempbox\mytemplen
    \mytemplen\dp\mytempbox
    \advance\mytemplen #2\relax
    \dp\mytempbox\mytemplen
    \colorbox{mygrey}{\hspace{1em}\usebox{\mytempbox}\hspace{1em}}}

\makeatother

\theoremstyle{definition}

\newcommand{\remove}[1]{}

\usepackage[textsize=tiny]{todonotes}

\begin{document}
\twocolumn[
\icmltitle{Efficient and Effective Methods for Mixed Precision Neural Network Quantization for Faster, Energy-efficient Inference}



\icmlsetsymbol{equal}{*}

\begin{icmlauthorlist}
\icmlauthor{Deepika Bablani}{equal,comp}
\icmlauthor{Jeffrey L. Mckinstry}{equal,comp}
\icmlauthor{Steven K. Esser}{comp}
\icmlauthor{Rathinakumar Appuswamy}{comp}
\icmlauthor{Dharmendra S. Modha}{comp}
\end{icmlauthorlist}

\icmlaffiliation{comp}{IBM Research, San Jose, California, USA}

\icmlcorrespondingauthor{Deepika Bablani}{deepika.bablani@ibm.com}
\icmlkeywords{Machine Learning, ICML}

\vskip 0.3in
]


\printAffiliationsAndNotice{\icmlEqualContribution} % otherwise use the standard text.

\begin{abstract}
   For efficient neural network inference, it is desirable to achieve state-of-the-art accuracy with the simplest networks requiring the least computation, memory, and power. Quantizing networks to lower precision is a powerful technique for simplifying networks. As each layer of a network may have different sensitivity to quantization, mixed precision quantization methods selectively tune the precision of individual layers to achieve a minimum drop in task performance (e.g., accuracy). To estimate the impact of layer precision choice on task performance, two methods are introduced: i) Entropy Approximation Guided Layer selection (EAGL) is fast and uses the entropy of the weight distribution, and ii) Accuracy-aware Layer Precision Selection (ALPS) is straightforward and relies on single epoch fine-tuning after layer precision reduction. Using EAGL and ALPS for layer precision selection, full-precision accuracy is recovered with a mix of 4-bit and 2-bit layers for ResNet-50,  ResNet-101 and BERT-base transformer networks, demonstrating enhanced performance across the entire accuracy-throughput frontier. The techniques demonstrate better performance than existing techniques in several commensurate comparisons. Notably, this is accomplished with significantly lesser computational time required to reach a solution.
\end{abstract}

\section{Introduction}
%\footnote{Preliminary work. Do not distribute.}
\begin{figure}[ht]
\vskip 0.2in
\begin{center}
\centerline{\includegraphics[width=\columnwidth]{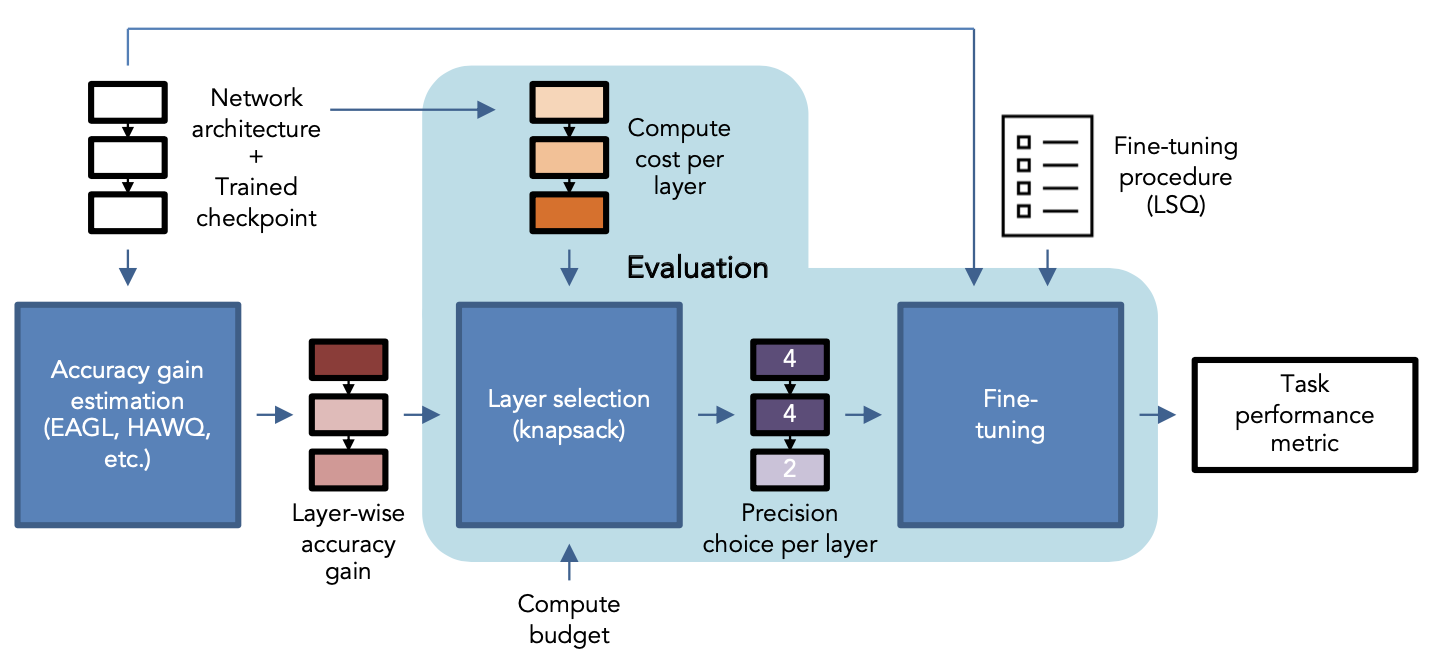}}
  \caption{\textbf{Evaluation framework for comparing layer precision selection approaches.} For a given network, computation budget, and fine-tuning procedure, identify the mixed precision method that provides the choice of precision per layer that achieves highest performance (e.g,. accuracy) on some task. A method under evaluation provides a layer-wise accuracy gain estimate, which is used along with corresponding computation costs by an optimization process to provide a precision choice per layer. The resulting network is then fine-tuned to provide performance on the task, which is used to rank the mixed precision methods considered.}
  \label{fig:flowchart}
\end{center}
\vskip -0.2in
\vspace{-1.7em}
\end{figure}
\vspace{1em}
Deep neural network quantization has emerged as an effective technique to improve network compression, throughput, and energy efficiency. Quantized networks use lower precision fixed point representation to approximate weights and activations, allowing them to operate on specialized hardware. This reduces model size and increases the number of operations in a given silicon area, enhancing parallelism for faster inference and energy efficiency. It is generally desirable to quantize as aggressively as possible without incurring accuracy degradation. Using the same uniform bit-width for most network layers can maintain accuracy at higher bit widths, but using extremely low precision still leads to significant accuracy degradation~\cite{courbariaux2015binaryconnect,esser2015backpropagation,rastegari2016xnor,zhou2016dorefa,mckinstry2019discovering,esser2020learned,liu2021nonuniform}. Mixed precision quantization methods aim to optimize each network layer's bit-width independently to maximize overall task performance while minimizing memory footprint and maximizing throughput~\cite{dong2019hawq,dong2020hawq,yao2021hawq,chen2021towards}. When these networks are implemented on specialized, highly efficient inference accelerators like the NorthPole Inference Processor~\cite{modha2023neural}, capable of performing 2-, 4-, and 8-bit operations efficiently, they offer substantial speed and power efficiency benefits in real-world applications.

Mixed precision quantization is challenging as the search space is prohibitively large. For a network with $L$ layers and $n$ precision choices per layer, $n^L$ different network configurations exist, making brute force precision selection for each layer impractical. Various methods are proposed in literature to mitigate this by using metrics that try to estimate the impact of quantization of individual layers on overall task performance ~\cite{dong2019hawq, dong2020hawq, yao2021hawq, chen2021towards, liu2021sharpness}, but they often target different criteria to optimize for, e.g. memory footprint or throughput, and fine-tune the resulting network with diverse algorithms. Hence, establishing a principled, direct comparison across mixed precision layer selection metrics based on reported results is not straightforward. The computational cost of many of these methods, combined with their reported inability to recover full precision accuracy at lower precisions (see Table~\ref{published} and Table~\ref{bert} for results on the ResNet-50~\cite{he2016deep} benchmark for image classification~\cite{deng2009imagenet} and the BERT-base~\cite{devlin2018bert} benchmark for natural language processing), underscore the need for i) new algorithms to solve this problem, and ii) a unifying framework for commensurate comparison.

Here, we introduce \textit{Accuracy-aware Layer Precision Selection (ALPS)} and \textit{Entropy Approximation Guided Layer selection (EAGL)}, two efficient and effective methods for choosing the bit-width configuration for network layers. Using these methods, we demonstrate networks capable of recovering full precision accuracy using a mix of 4-bit and 2-bit layers for ResNet-50 and ResNet-101 for image classification~\cite{deng2009imagenet} and BERT-base transformer network for natural language question answering~\cite{rajpurkar2016squad}. These mixed precision networks offer significant improvements in energy-efficiency over their fixed precision counterparts. For quantitative measures of improvement in energy-efficiency, we refer the reader to Fig. S5~\cite{modha2023neural} for implementations of the solutions found using these techniques on NorthPole and measured inference power numbers. The techniques proposed here are briefly described next.
\begin{table}[ht!]
%\centering
\caption{\textbf{EAGL and ALPS demonstrate state-of-the-art results for classification~\cite{deng2009imagenet} with ResNet-50 compared to techniques from literature}. Top-1 Accuracy Drop is the accuracy gap between the full precision (FP32) baseline and the low precision (LP)/mixed precision (MP) network. Negative values indicate that the MP network performs better than the FP32 network. Numbers in bracket indicate FP32 and MP scores, respectively. Compression Ratio is model compression w.r.t FP32 weights. BOPS is Giga-Bit Operations~\cite{yao2021hawq}. HAQ and LQ-Nets~\cite{zhang2018lq} keep activations at FP32. PACT~\cite{choi2018pact} and RVQuant~\cite{park2018value} do not used mxed precision. "-" indicates data not published. FP32 accuracy is different for each method as authors use different checkpoints as the starting point. This highlights a need for a commensurate comparison framework, as an apples-to-apples comparison is challenging.}
\label{published}
\vskip 0.15in
\begin{center}
\begin{small}
\begin{sc}
\begin{tabular}{lcccr}
\toprule
Method & \begin{tabular}[c]{@{}c@{}}Top 1 \\ Accuracy \\Drop\end{tabular} & \begin{tabular}[c]{@{}c@{}}Compression\\ Ratio\end{tabular} &  BOPS\\
\midrule
\begin{tabular}[c]{@{}c@{}}ALPS\\  (Ours)\end{tabular}  & \textbf{\begin{tabular}[c]{@{}c@{}}-0.22 \\ (76.13-76.35)\end{tabular}}&  10.31$\times$ & \textbf{34}\\ \hline
\begin{tabular}[c]{@{}c@{}}EAGL\\  (Ours)\end{tabular}  & \textbf{\begin{tabular}[c]{@{}c@{}}-0.17 \\ (76.13-76.30)\end{tabular}}&  10.31$\times$ & \textbf{34}\\ \hline
HAQ&\begin{tabular}[c]{@{}c@{}}0.67 \\ (76.15 - 75.48)\end{tabular}& 10.57$\times$ &520   \\ \hline
\citeauthor{chen2021towards}
&\begin{tabular}[c]{@{}c@{}}0.85 \\ (76.13-75.28)\end{tabular}& 12.24$\times$ & -  \\ \hline
HAWQ-v3&\begin{tabular}[c]{@{}c@{}}0.99 \\ (77.72-76.73)\end{tabular}& - &154   \\ \hline
HAWQ-v2&\begin{tabular}[c]{@{}c@{}}1.63 \\ (77.39-75.76)\end{tabular}& 12.24$\times$&  - \\ \hline
HAWQ&\begin{tabular}[c]{@{}c@{}}1.91 \\ (77.39-75.48)\end{tabular}& 12.28$\times$ & -\\ \hline
AutoQ&\begin{tabular}[c]{@{}c@{}}2.29 \\ (74.80-72.51)\end{tabular}& 10.26$\times$  &- \\ \hline
SPSS&\begin{tabular}[c]{@{}c@{}}0.93 \\ (77.15-76.22)\end{tabular}& 8.00$\times$ & -  \\ \hline
BP-NAS&\begin{tabular}[c]{@{}c@{}}1.85 \\ (77.56-75.71)\end{tabular}& 11.03$\times$ & -  \\\hline
\begin{tabular}[c]{@{}c@{}}OMPQ\\ \citeauthor{ma2023ompq}\end{tabular} &\begin{tabular}[c]{@{}c@{}}1.52 \\(77.72-76.20)\end{tabular}&- &141\\ \hline
PACT&\begin{tabular}[c]{@{}c@{}}0.20\\(76.90-76.70)\end{tabular}&- &101\\ \hline
RVQuant&\begin{tabular}[c]{@{}c@{}}0.32\\ (75.92-75.60)\end{tabular}&- &101\\ \hline
LQ-Nets&\begin{tabular}[c]{@{}c@{}}1.32\\(77.72-76.40)\end{tabular}&- &486\\ 
\bottomrule
\end{tabular}
\end{sc}
\end{small}
\end{center}
\vskip -0.1in
\end{table}
%\end{tabular}
%\end{table}

\begin{table}[ht!]
%\centering
\caption{\textbf{EAGL and ALPS demonstrate state-of-the-art results for natural language question answering with BERT-base compared to techniques from literature}. Most techniques do not quantize activations below 8-bits and all techniques use different training and fine-tuning recipes, so comparisons are not always apples-to-apples, but EAGL and ALPS are able to find solutions that exceed full precision accuracy at less than 4-bits on average, better than all published results.W-bits and A-bits represents weight and activation bits used throughout the network. 4/2 implies a mix of 4-bit and 2-bit layers in the model. F1 score is the task metric for question answering and the drop (in brackets) is the accuracy gap between the full precision (FP32) baseline and the mixed precision (MP) network. Negative values indicate that the MP network performs better than the FP32 network. Compression Ratio is model compression w.r.t FP32 weights.}
\label{bert}
\vskip 0.15in
\begin{center}
\begin{small}
\begin{sc}
\begin{tabular}{lcccr}
\toprule
Method & W-Bits & A-Bits & \begin{tabular}[c]{@{}c@{}}F1 \\(Drop)\end{tabular} & \begin{tabular}[c]{@{}c@{}}Comp-\\ ression\\Ratio\end{tabular} \\
\midrule
\begin{tabular}[c]{@{}c@{}}EAGL\\  (Ours)\end{tabular}  & 4/2 & 4/2 & \textbf{\begin{tabular}[c]{@{}c@{}}88.29 \\ (-0.09)\end{tabular}}& 8.40$\times$ \\ \hline
\begin{tabular}[c]{@{}c@{}}ALPS\\  (Ours)\end{tabular}  & 4/2 & 4/2 & \textbf{\begin{tabular}[c]{@{}c@{}}88.26 \\ (-0.06)\end{tabular}}&  8.74$\times$ \\ \hline
\begin{tabular}[c]{@{}c@{}}Ternary\\  BERT\\\citeauthor{zhang2020ternarybert}\end{tabular}& 2 & 8 &\begin{tabular}[c]{@{}c@{}}87.67 \\ (1.02)\end{tabular}& 14.90$\times$ \\ \hline
\begin{tabular}[c]{@{}c@{}}Q8BERT\\\citeauthor{zafrir2019q8bert}\end{tabular} & 6 & 8 &\begin{tabular}[c]{@{}c@{}}85.48 \\ (3.21)\end{tabular}& 5.20$\times$ \\ \hline
Q8BERT & 4 & 8 &\begin{tabular}[c]{@{}c@{}}23.82 \\ (64.87)\end{tabular}& 8.00$\times$ \\ \hline
\begin{tabular}[c]{@{}c@{}}KDLSQ-\\  BERT\\\citeauthor{jin2021kdlsq}\end{tabular}& 4 & 8 &\begin{tabular}[c]{@{}c@{}}89.21 \\ (-0.52)\end{tabular}& 7.70$\times$ \\ \hline
\begin{tabular}[c]{@{}c@{}}KDLSQ-\\  BERT\end{tabular}& 2 & 8 &\begin{tabular}[c]{@{}c@{}}88.45 \\ (0.24)\end{tabular}& 14.90$\times$ \\ \hline
\citeauthor{wang2022deep}& 8 & 8 & \begin{tabular}[c]{@{}c@{}}88.35 \\ (0.34)\end{tabular}& 4.00$\times$\\\hline
\citeauthor{wang2022deep}& 4 & 4 & \begin{tabular}[c]{@{}c@{}}87.86 \\ (0.83)\end{tabular}& 8.00$\times$\\
\bottomrule
\end{tabular}
\end{sc}
\end{small}
\end{center}
\vskip -0.1in
\end{table}
%\end{tabular}
%\end{table}

The most straightforward way to estimate the contribution of each layer to overall network accuracy, assuming independent contributions, is to lower the precision of each layer one at a time, fine-tune briefly, e.g. for one epoch, and use this accuracy difference as a measure of the relative advantage of keeping a layer at high precision. Intuitively, the network accuracy can be optimized by keeping the layers which show the largest difference in accuracy at higher precision, and choosing the layers with low difference in accuracy for further quantization. This method is referred to as Accuracy-aware Layer Precision Selection (ALPS). Given a fixed dataset, the computational complexity is proportional to the number of layers.

EAGL, on the other hand, is built upon the insight that the complexity required by a given layer to achieve good performance can be estimated using the entropy of the empirical distribution of its parameters after training, assuming appropriate regularization. Intuitively, the advantage of keeping a layer at a higher precision is expected to be directly proportional to the entropy of the empirical weight distribution post training. For example, a layer that has entropy close to 4-bits should be kept at 4-bits, while a layer with entropy close to 2-bits is a good candidate for further quantization. This formulation relies on empirical estimates for the marginal distributions of layer parameters, and thus, given a trained network, EAGL is a computationally inexpensive technique that does not require access to the training dataset, and, surprisingly, can perform as well as the more direct accuracy aware method described above.

Once the layer-wise accuracy gain estimates are determined as described above (using ALPS, EAGL, or another mixed precision technique from the literature), along with the known computational cost of each layer, the next step is to select the precision of each layer from two or more precision choices in order to maximize the estimated task performance of the resulting network without exceeding a target computational budget. In the experiments presented in this paper, we limit the search space for each layer to two precisions, 4-bit and 2-bit. Both weights and activations in a layer are quantized to the chosen precision. We solve this optimization problem efficiently by formulating it as a 0-1 Integer Knapsack Problem from combinatorial optimization~\cite{martello1990knapsack}. Finally, the layer precision choices from the optimization step are used to create a mixed precision network which is fine-tuned until convergence to get the final task performance. This process is depicted in Figure~\ref{fig:flowchart}. Using this evaluation framework, ALPS and EAGL are demonstrated to outperform leading mixed precision layer selection techniques like HAWQ-v3~\cite{yao2021hawq} and achieve state-of-the-art accuracy using ResNet-50 and ResNet-101 networks for image classification~\cite{deng2009imagenet}, PSPNet~\cite{zhao2017pyramid} for semantic segmentation~\cite{cordts2016cityscapes} and BERT-Base for natural language question answering~\cite{rajpurkar2016squad} using 2-bit and 4-bit layer mixed precision networks. 

The main contributions of this paper are (i) two elegant measures for estimating the impact of layer precision on task performance that achieve state-of-the-art results for mixed precision neural networks, and (ii) a framework for fair comparison across varying approaches that decomposes the bit-width configuration selection problem into three steps - accuracy gain estimation, layer-wise precision selection for a given budget using 0-1 Integer Knapsack optimization algorithm, and  fine-tuning of resulting mixed precision network.
\section{Related Work} \label{Sec:modelling_related}
%\paragraph{Quantization and learning bit-width configuration}
Methods for optimizing the layer-wise precision of deep neural networks can be divided into several categories. Network Architecture Search methods can naturally incorporate precision as an additional variable but are generally computationally expensive~\cite{wu2018mixed,chen2018joint,yu2020search,sun2021effective}. HAQ~\cite{wang2019haq} and AutoQ~\cite{lou2019autoq}, use reinforcement learning (RL) in the search process. A recent technique~\cite{liu2021sharpness} uses RL in a more efficient manner and employs a separate network trained with RL which learns the distribution of precisions which work well as the network being optimized is trained with random, layer-wise precisions.
There are also methods which learn the precision of each layer using stochastic gradient descent (SGD) with a modified loss function~\cite{nikolic2020bitpruning,yang2020fracbits}. Hessian Aware methods, like HAWQ~\cite{dong2019hawq} and its successors, HAWQ-v2~\cite{dong2020hawq}, HAWQ-v3~\cite{yao2021hawq}, and Chen et al.~\cite{chen2021towards} evaluate the curvature of the loss function for each layer to predict, given the quantization error of that layer, how much quantization will increase the loss. HAWQ-v3 achieves state-of-the-art in network compression and throughput. A recent accuracy aware method~\cite{liu2021layer} tries to estimate the accuracy gain for each layer using imprinting, which uses one epoch of fine-tuning per layer. Although the authors use insights from few-shot learning to speed up the layer-wise training, a drawback of this method is that it requires network surgery, removing downstream layers and replacing them with a single linear layer, thus evaluating each layer out of context.
\section{Methods}
Mixed precision quantization is based on the intuition that different layers in a network have different sensitivity to quantization and hence some layers are better candidates for more aggressive quantization than others. In particular, we are interested in solving the following problem.
\paragraph{Problem formulation}Given a choice between two precisions, $b_1$ and $b_2$ for each of the layers in a network, find the bit-width configuration for all the layers that achieves the best task performance while remaining within a given computational budget for the network. 

The budget can be any target metric evaluating performance on a hardware inference platform, for instance, the memory footprint, latency, power, or a combination of multiple metrics. While the problem is formulated as a binary choice problem above, where the precision of a layer can be one of two precision choices, $b_1$ or $b_2$, this can be extended to the case of more than two precision choices as well. A model is considered for which a layer's computational cost is linear in its bit-width, but the techniques are equally applicable to other cost models such as quadratic.

The evaluation methodology, outlined in the next sub-section, provides a unifying framework for commensurate comparison between various popular mixed precision techniques from literature, as well as the novel layer selection methods introduced in this work. As emphasized earlier, there is a need for a unifying framework for evaluating techniques for mixed precision layer selection, as the lack of an apples-to-apples comparison between different methods proposed in literature and different objectives being optimized for, each with its own merits and limitations, makes it very challenging for a practitioner to select a candidate that is the most suitable for their problem of interest due to the absence of a coherent comparison methodology. It is hoped that the framework for comparison proposed here serves as a benchmark for evaluation of future methods as well. \remove{and we will be releasing our code to further aid this process.}

\subsection{Evaluation Methodology} \label{knapsack}
The proposed evaluation framework (Figure \ref{fig:flowchart}) is parameterized by a task, a network architecture, a target computational budget, and a fine-tuning recipe. It takes as input a per layer scalar measure of the expected relative task performance improvement when layer $l$ is quantized at precision $b_1$ instead of a lower precision $b_2$, which is referred to as accuracy gain, $G_l$. As an example, this is the accuracy gained by quantizing a layer in a classification network at 4-bit instead of 2-bit. The higher the gain, the more desirable it is to keep this layer precision at 4-bit instead of 2-bit. Different methods for mixed precision quantization propose different ways of quantifying this cost per layer. This is the key distinction between techniques evaluated in this framework. 

Once the layer-wise gain estimates are obtained, the optimization problem of precision selection of the layers of a model is formulated formally as follows:

Maximize $\sum_{l=1} ^{L} G_l P_l$, i.e., the total accuracy gained while keeping $\sum_{l=1} ^{L} C_l \leq B$, where $L$ is the number of layers, $G_l$ is the accuracy gain per layer as defined above, $P_l$ is 1 if the precision of layer $l$ is kept at the higher precision $b_1$, and 0 otherwise, $C_l$ is the computational cost of layer $l$ in cycles, and $B$ is the overall computational budget in cycles. 

This optimization problem is formulated as a 0-1 Integer Knapsack Problem from combinatorial optimization, which is defined next. For a given a set of N distinct items, each with integer value, $V_i$, and integer weight, $W_i$, find the subset of items to put into a knapsack with integer capacity, $C$, which maximizes the sum of the values of the items placed into the knapsack, such that the sum of the weights of those items does not exceed $C$.  "0-1" here refers to the fact that each item can be either in or out of the knapsack.  

The Layer Selection problem is mapped onto the 0-1 Integer Knapsack problem as follows. The $L$ layers of the network are the $N$ items that can be selected for inclusion in the knapsack, where including the layer $l$ means choosing the higher of the two precision choices for the layer. More concretely, $G_l$, an estimate of the accuracy gained by keeping $l$ at a higher precision $b_1$ instead of quantizing it to a lower precision $b_2$, is the value of each item. Each mixed precision method under consideration assigns $G_l$ for each of the layers. The computational cost difference between keeping the layer at $b_1$ instead of $b_2$ is the weight of each item in the knapsack setup. Finally, the total computational budget, $B$, of the network is the knapsack capacity, $C$. It is possible to optimize for a different target of interest like memory footprint or latency.\footnote{This formulation is based on the assumption that the contributions of individual layers to overall task accuracy are additive. See Appendix \ref{Sec:additive} for experiments demonstrating the additive nature of layer-wise accuracy contributions.}
\footnote{The Knapsack formulation provides us with an \(\epsilon\)-optimal solution in terms of the granularity of the value estimates. The floating point estimates from any method under evaluation for $G_l$ for each layer are quantized to integer values between 1 and 10000, limiting errors to \(10^{-5}\).}

Chen et al.~\cite{chen2021towards} formulate the mixed precision layer selection problem as a general knapsack problem, but given that only two precision choices are considered here (2-bit and 4-bit), the further simplification of mapping to the 0-1 Integer Knapsack problem allows efficient solutions in $\mathcal{O}(BL)$~\cite{martello1990knapsack}. The Python implementation of the knapsack solver takes 2.3 seconds for ResNet-50, 3.5 seconds for ResNet-101, and 1 minute, 18 seconds for PSPNet. With the choice of precision for each layer from the optimizer, a mixed precision network is created with this configuration, which is then fine-tuned, to get the task performance of the mixed precision network chosen by the technique. The resulting task performance is the final output of the evaluation. 

For comparison across different mixed precision techniques, each method provides an accuracy gain, $G_l$ for each layer of a network, and then follows the above standardized evaluation protocol. For all of the experiments in this paper, 4-bit and 2-bit mixed precision choices are used in the problem setup because either previous published works have already demonstrated full precision accuracy for the networks under consideration at 4-bit ~\cite{mckinstry2019discovering, esser2020learned}, or fine-tuning with the proposed training recipe was sufficient to recover the task performance of a full precision network using a 4-bit network. In all experiments, by sampling budgets in between the throughput of a 4-bit and a 2-bit network, multiple points along the throughput-accuracy frontiers were sampled for each technique, thereby demonstrating a very compelling test framework to compare different techniques across a wide variety of budgets in an unbiased manner.

The next two subsections describe the two methods introduced here for the problem of mixed precision layer selection described above -- specifically the key sub-problem of accurately estimating accuracy gains, $G_l$ for each network layer $l$ -- followed by implementation details specific to the evaluation framework. In the following section, the techniques are compared with state-of-the-art mixed precision techniques from literature and demonstrate superior performance, not only for a fixed budget, but along the entire frontiers.

\subsection{Accuracy-aware Layer Precision Selection for quantization (ALPS)}\label{accaware}
Most published approaches that propose solutions to the  mixed precision layer selection problem are indirect proxies for what is really needed -- an estimate for the accuracy contribution of each layer \emph{after} quantization aware training~\cite{yao2021hawq}. In~\cite{liu2021sharpness}, the authors do measure the accuracy of a fine-tuned network, but indirectly using a second network. Here, we propose an accuracy aware method which directly measures how well each layer can be fine-tuned. The intuition is that layers that have the highest accuracy gain when quantized to a higher precision $b_1$ (here, 4-bit) instead of a lower precision $b_2$ (here, 2-bit) are good candidates to be kept at $b_1$ in a mixed precision network. This approach is an intuitive, simpler variant of ~\cite{liu2021layer}. 

\begin{algorithm}[tb]
    \caption{ALPS for layer selection}
    \label{alg:alps}
\begin{algorithmic}
\REQUIRE Pre-trained ${L}$-layer model $M$ at precision $b_1$ (4-bit)
\FOR{\texttt{$l \gets 1$ $\mathbf{to}$ $L$}} 
    \STATE {Drop layer $l$'s precision to $b_2$ (2-bit), all others remain at $b_1$}
    \STATE {Fine-tune resulting network for 1 epoch} 
    \STATE {$A_l, Loss_l \gets $ final training set Accuracy, Loss}
\ENDFOR 
\IF{$M = $ PSPNet}
    \STATE {$G_l \gets Loss_l$ for all $l$ } 
\ELSE
    \IF{$M =$ ResNet}
    \STATE {$G_l \gets max(A) - A_l$ for all $l$ } 
    \COMMENT{Accuracy gained using precision $b_1$ (4-bit) instead of $b_2$ (2-bit)}
    \ENDIF
\ENDIF
\end{algorithmic}
\end{algorithm}

Algorithm~\ref{alg:alps} describes ALPS. For a given network, the default training parameters used for training the higher precision (here, 4-bit) model are used. Starting from a fully trained 4-bit model, the precision of each convolution layer is lowered from 4-bit to 2-bit, one at time, and the resulting network is fine-tuned (with all layers at 4-bits and a single layer at 2-bit) for the equivalent of one training epoch (1 epoch for ResNet-50 and ResNet-101 and 250 iterations for PSPNet). The average training set performance (accuracy and loss for Resnet and PSPNet models, respectively) over the training period is then used to estimate the accuracy gained, $G_l$ if the given layer remained at 4-bit. The assumption made here, that total network accuracy is the sum of the layer-wise accuracies, is justified theoretically ~\cite{zhou2018adaptive}, and empirically in the Appendix \ref{Sec:additive}. To the extent that this linearity holds, ALPS avoids the need to try all combinations of layer precision settings. 
These measures are provided to the optimizer, along with the desired computational cost target, or budget, in order to choose a list of precisions for the network layers which maximizes the overall task performance of the network while satisfying the budget constraint. 

\subsection{Entropy Approximation Guided Layer selection for quantization (EAGL)}\label{EAGL}
EAGL uses empirical entropy to find a solution for the problem formulation above. The key insight in the development of this metric is that the entropy of the empirical distribution (Eq.~\eqref{eqn:probability}) of parameters of a layer in a network represents a measure of the required complexity to achieve the desired performance. With this insight, we develop an entropy-based metric for quantifying the advantage of keeping a layer at a higher precision. 

\begin{figure}[ht]
\vskip 0.2in
\begin{center}
\centerline{\includegraphics[width=\columnwidth]{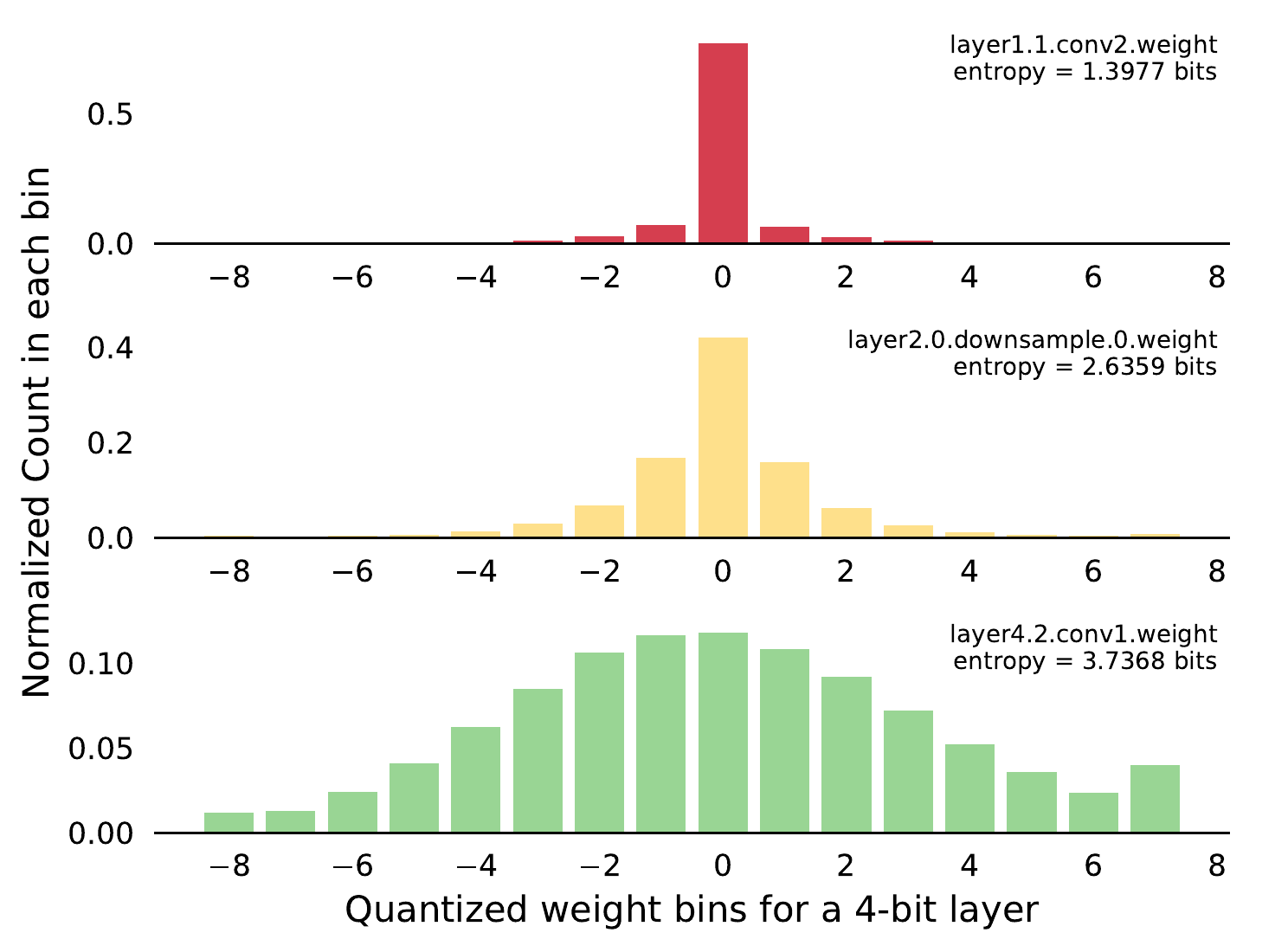}}
 \caption{\textbf{Histogram of normalized counts of quantized weights in each bin for 3 layers of a trained 4-bit ResNet-101 network.} EAGL predicts that layers with lower entropy are better candidates for further quantization. For example, between the three layers shown above, EAGL predicts that quantizing the first layer (entropy $= 1.3977$ bits) to 2 bits has lower impact on task accuracy than quantizing the third layer (entropy $= 3.7368$ bits).}
  \label{fig:entropy}
\end{center}
\vskip -0.2in
\vspace{-1.5em}
\end{figure}

Consider a layer $l$ in a network at precision $b$. Post quantization, a parameter in $l$ can have a value equal to one of $2^b$ distinct values, i.e., the weight can occupy one of $2^b$ bins. When parameters of a layer are spread out evenly in the available bins, and the distribution approaches a uniform distribution, its entropy is high. However, if many of the weights are concentrated in a small subset of bins, the entropy of the distribution of quantized weights is low. Intuitively, a layer with a lower entropy is a better candidate for further quantization as the representation of the transformation learned by the layer parameters can be compressed further more easily. This is illustrated in Fig. \ref{fig:entropy} using 3 convolutional layers from a trained 4-bit ResNet-101 network. 

Let $\hat{p}^b$ be the empirical distribution of the layer's quantized parameters at precision $b$. EAGL is based on the hypothesis that $H(\hat{p}^b)$\footnote{$H$ is used everywhere to denote the discrete entropy function.} provides a good measure which can serve as the accuracy gain metric in the mixed precision problem formulation described earlier. Intuitively, $H(\hat{p}^b)$ represents the number of bits needed to represent the parameters in that layer according to the empirical distribution, as opposed to the actual allocated bits $b$. If $H(\hat{p}^b)$ is close to the allocated bit-width $b$, the layer is a bad choice for further quantization; however, if $H(\hat{p}^b)$ is much lower than $b$, the layer is a better candidate for further quantization to a lower precision. This metric can be calculated  practically as follows.

Let the vector $\mathbf{w}$ represent $n$ trained parameters for layer $l$, and denote $\mathbf{w}=$ $(w_1,w_w,\ldots,w_n)$ where $w_i$ are quantized with bit-width $b$. If $w_i$ takes values from a discrete set $\mathcal{A}$, then $|\mathcal{A}| \le 2^{b}$. For $c \in \mathcal{A}$, the empirical  probability distribution is computed by

\begin{equation}
\label{eqn:probability}
\hat{p}^b_{l} (c) = \frac{1}{n}  \sum_{i=1}^{n} \mathbbm {1} _{c} (w_i), 
\end{equation}
%\begin{minipage}{.1\linewidth} 
\text{where}
%\end{minipage}%
%\begin{minipage}{.9\linewidth} 
\begin{equation}
\label{eqn:indicator}
\mathbbm {1} _{c}(w_i):=
\begin{cases}
1 &\text{if}  \, w_i  = c\\
0 &\text{if}  \, \text{otherwise}. 
\end{cases}
\end{equation}
%\end{minipage}%
%\begin{minipage}{.35\linewidth} 

%\end{minipage} \\
%
Using $\hat{p}^b_{l}$ from equation \ref{eqn:probability}, we compute its entropy as
\begin{equation}
\label{eqn:entropy}
H(\hat{p}^b_{l}) = -\sum_{c} \hat{p}^b_{l} (c) \, \log{\hat{p}^b_{l} (c) }.
\end{equation}

The value of keeping layer $l$ at higher precision $b$ is quantified as the value of this entropy measure $H(\hat{p}^b_{l})$(Eq.~\eqref{eqn:entropy}). 

Algorithm~\ref{alg:cap} describes EAGL. EAGL requires only one pre-trained model -- one where all layers are quantized at precision $b$. The empirical distribution of parameters is estimated for each of the layers in the network, and then  $H(\hat{p}^b_{l})$ is calculated for each layer. These layer-wise measures are provided to the optimizer in the context of our full evaluation framework. EAGL is (i) easy to approximate, and (ii) does not need access to the training dataset to compute, making it faster and more generally applicable to other problem domains than the other metrics compared against.
\begin{algorithm}[tb]
    \caption{EAGL for layer selection}
    \label{alg:cap}
\begin{algorithmic}
\REQUIRE Pre-trained ${L}$-layer model at precision $b$
%,  (iii) compute budget $B$, and (iv) for each layer $l$, additional compute cost $C_l$ when using bit-with $b_1$ compared to using $b_2$ for that layer.
\FOR{\texttt{$l \gets 1$ $\mathbf{to}$ $L$}} 
    \STATE{$G_l \gets H(\hat{p}^{b}_{l})$} 
    \COMMENT{Calculate $G_l$ for each layer using ${H}$  defined in equation \ref{eqn:entropy}}
    %\State {$G_l \gets \max(0,{{H}(\bhi{\hat{p}}_{W_l}) - %{H}(\blo{\hat{p}}_{W_l})})$} %\Comment{${H}$  defined in equation %\ref{eqn:entropy}}
\ENDFOR
\end{algorithmic}
\end{algorithm}

This work makes use of the Minimal Description Length (MDL) principle~\cite{rissanen1978modeling}, and its application, in particular, to learning the required precision for representing network parameters~\cite{ Wallace1990ClassificationBM,Hinton1993KeepingTN, hochreiter1997flat,ullrich2017soft,blier2018description,wiedemann2019entropy}. One approach is related to attempts to build MDL principle into the optimization function ~\cite{Wallace1990ClassificationBM,Hinton1993KeepingTN}, and derive a learning rule to minimize the number of bits needed to represent the weights (also see ~\cite{wiedemann2019entropy} and references therein). Here, on the other hand, a slightly different approach is taken, and simple insights from information theoretic principles are used to propose a measure that uses the entropy of a layer's learned parameters as the metric to choose between layers to achieve the most profitable trade-off between bits allocated to represent layer parameters and network performance. 

A probability distribution on model parameters $W$ is assumed, and $\hat{p}^b_l$ is used as an estimate of the true marginal distribution. This estimation is valid when the layer parameters are assumed to be independent and identically distributed (iid). Even if the iid assumption is not satisfied, it  may be assumed that the parameters in each of the layers are identically distributed and that they satisfy a form of the strong-mixing ~\cite{pollicott1998dynamical} condition, or equivalently that the parameters that are sufficiently far apart are independent. 
This practical assumption allows the use of a single instance of parameters learned using SGD and binning to estimate the distribution of these parameters and allows the efficient computation of the entropy terms. It is emphasized that no change to the loss function was made to minimize the measured entropy and that SGD with regularization is relied upon entirely to trade-off between entropy of the empirical distribution and network performance for a given precision. 

\subsection{Implementation details}
The implementation details and assumptions made throughout the paper are described next. 

\subsubsection{Setting layer precision}
The precision of a layer dictates how its input activations and weights are represented, and therefore, the precision of the input activations and weights must match for a given layer. If an activation tensor provides input to multiple layers, all such layers must therefore have the same weight precision as well. Such layers are considered linked, and are treated as a single layer with a computational cost and accuracy gain measure that is the summation of its member values. All other data is represented with, and operations occur at, full precision. A unit of Bit Multiply-Accumulate operations (BMACs) is used for computational cost calculation, which is defined as BMAC = $b$*MAC, where $b$ is the precision of the layer (weights and activations) in bits. The first and last layers are fixed at 8-bit, following a common practice for quantized networks ~\cite{zhou2016dorefa, zhu2016trained}, and intermediate layers with less than 128 input features are fixed at 4-bit. Layers with fixed precision do not count towards the computational budget. 

\subsubsection{Task evaluation details}
Three distinct tasks are evaluated here, image classification~\cite{deng2009imagenet} using ResNet-50 and ResNet-101~\cite{he2016deep}, and semantic segmentation~\cite{cordts2016cityscapes} using PSPNet~\cite{zhao2017pyramid}, and natural language question-answering~\cite{rajpurkar2016squad} using BERT-base~\cite{devlin2018bert}. For all three tasks, instead of choosing a fixed budget apriori, a sweep of computational budgets is used as evaluation points. This ensures that the evaluation framework is fair and compares techniques across a suite of evaluation points. Some techniques may demonstrate stronger performance for tighter budget constraints while others may be better for less constrained budgets; and hence evaluating only on a single budget can potentially unfairly benefit one technique over another. Eight equally spaced computational budgets are used for ResNet networks and 4 equally spaced computational budgets for PSPNet between the computational cost of a 4-bit and a 2-bit network.

\subsubsection{Training and implementation details}
All models are fine-tuned with weights and activations quantized using LSQ ~\cite{esser2020learned}. 

The ResNet networks are adapted from the PyTorch Model Zoo, and trained for image classification~\cite{deng2009imagenet} with weight decay of 1e-4, initial learning rate of 0.01 with cosine learning rate decay ~\cite{loshchilov2016sgdr}, batch size 128, and using knowledge distillation ~\cite{hinton2015distilling}. $160\times160$ images are used to speed up training, while $224\times224$ images are used for testing. The initial quantization step-size for all layers being reduced from 4- to 2-bit is set to $4s$, where $s$ is the learned step-size loaded from the 4-bit checkpoint. Models at 4-bit, used as the initial checkpoint to fine-tune mixed precision models, are trained by fine-tuning the full precision checkpoint for 90 epochs. Each mixed precision model is then further fine-tuned for 90 epochs for comparison. 

PSPNet is adapted from the PyTorch Segmentation Toolbox~\cite{huang2018pytorch} and trained on semantic segmentation~\cite{cordts2016cityscapes} with an learning rate of 0.015, weight decay of 5e-5, and batch size 4. Models at 2-bit and 4-bit are trained by fine-tuning a full precision checkpoint for 80,000 and 10,000 iterations respectively, and mixed precision methods are fine-tuned from the 4-bit trained model for 40,000 iterations.  

For BERT, the 4-bit network used as the starting point for mixed precision runs is initialized with a full precision Huggingface checkpoint finetuned for Squad1.1\footnote{\url{https://huggingface.co/csarron/bert-base-uncased-squad-v1}}. For the mixed precision budget sweeps, we use a learning rate of 3.8e-3 with a batch size of 192 for the lamb optimizer [1] with knowledge distillation. For each technique, we use the same 4-bit checkpoint fine-tuned from a full precision checkpoint using FAQ for weight initialization. We scale the learned scale factors by 4 for the layers that a technique chooses to drop from 4-bit to 2-bit. This provide a good initial estimate for further training. 

\section{Results} \label{results}
The performance of ALPS and EAGL is compared against leading mixed precision techniques from the literature in Table~\ref{published} using the most commonly used benchmark, ResNet-50 for image classification~\cite{deng2009imagenet} and the BERT-base~\cite{devlin2018bert} benchmark for natural language processing. Both techniques have the highest accuracy (no accuracy drop) for the most efficient networks (lowest BOPs). As seen in the table, different techniques are quantized starting from different full precision accuracy checkpoints, trained using different training recipes, and report on different metrics. To mitigate these issues, HAWQ-v3, a leading mixed precision layer selection technique, was re-implemented and results are presented below from apples-to-apples comparison with HAWQ-v3. Other techniques could not be re-implemented due to the significant effort required to reproduce results without released code. Four distinct networks were evaluated on two large-scale computer vision and one natural language dataset, across various computational budgets, demonstrating strong evidence of the effectiveness of these methods.

\subsection{Evaluation using ResNet for image classification}  \label{Sec:resnetresults}
The performance of ALPS and EAGL for layer selection is evaluated on image classification~\cite{deng2009imagenet} using ResNet-50 and ResNet-101, using the methodology described in Section \ref{knapsack}. The methods are compared with a re-implementation of HAWQ-v3~\cite{yao2021hawq}\footnote{See Appendix \ref{Sec:hawqdetails} for details of the re-implementation of HAWQ-v3 for these experiments.} and three baselines -- a uniform cost baseline, where every layer is given the same value for staying at 4 bits, a method which ranks layers from first to last, and drops the precision of the first $n$ layers greedily until the resulting network just meets the budget, and a third method which ranks layers from last to first. Instead of choosing a fixed budget apriori, eight different target computational budgets are sampled for evaluation, roughly at 95\%, 90\%, 85\%, 80\%, 75\%, 70\%, 65\% and 60\% of the computational cost of a 4-bit network. For each budget, networks created using each technique compared are fine-tuned for 90 epochs using 5 seeds. 

\begin{table}[ht]
\caption{\textbf{Metric computational cost comparison for ResNet-50 and PSPNet.} The cost compared here includes only the computational resources required for the layer-wise metric estimation for each method, and not the cost for fine-tuning the resulting mixed precision network, nor the cost for training the initial 4-bit checkpoint, as these are the same for all techniques.}
\label{bakeoffR50}
\vskip 0.15in
\begin{center}
\begin{small}
\begin{sc}
\begin{tabular}{lcccr}
\toprule
Method & ResNet-50 & PSPNet \\ 
\midrule
\textbf{EAGL(ours)} & \textbf{3.15 CPU seconds} & \textbf{<1 CPU minute}\\ \hline 
ALPS(Ours) & 166 GPU hours & 67 GPU hours \\ \hline
HAWQ-v3 & 2 GPU hours & ~1032 GPU hours\\ 
\bottomrule
\end{tabular}
\end{sc}
\end{small}
\end{center}
\vskip -0.1in
\end{table}

\begin{figure*}[ht]
\vskip 0.15in
\begin{center}
\includegraphics[width=\columnwidth]{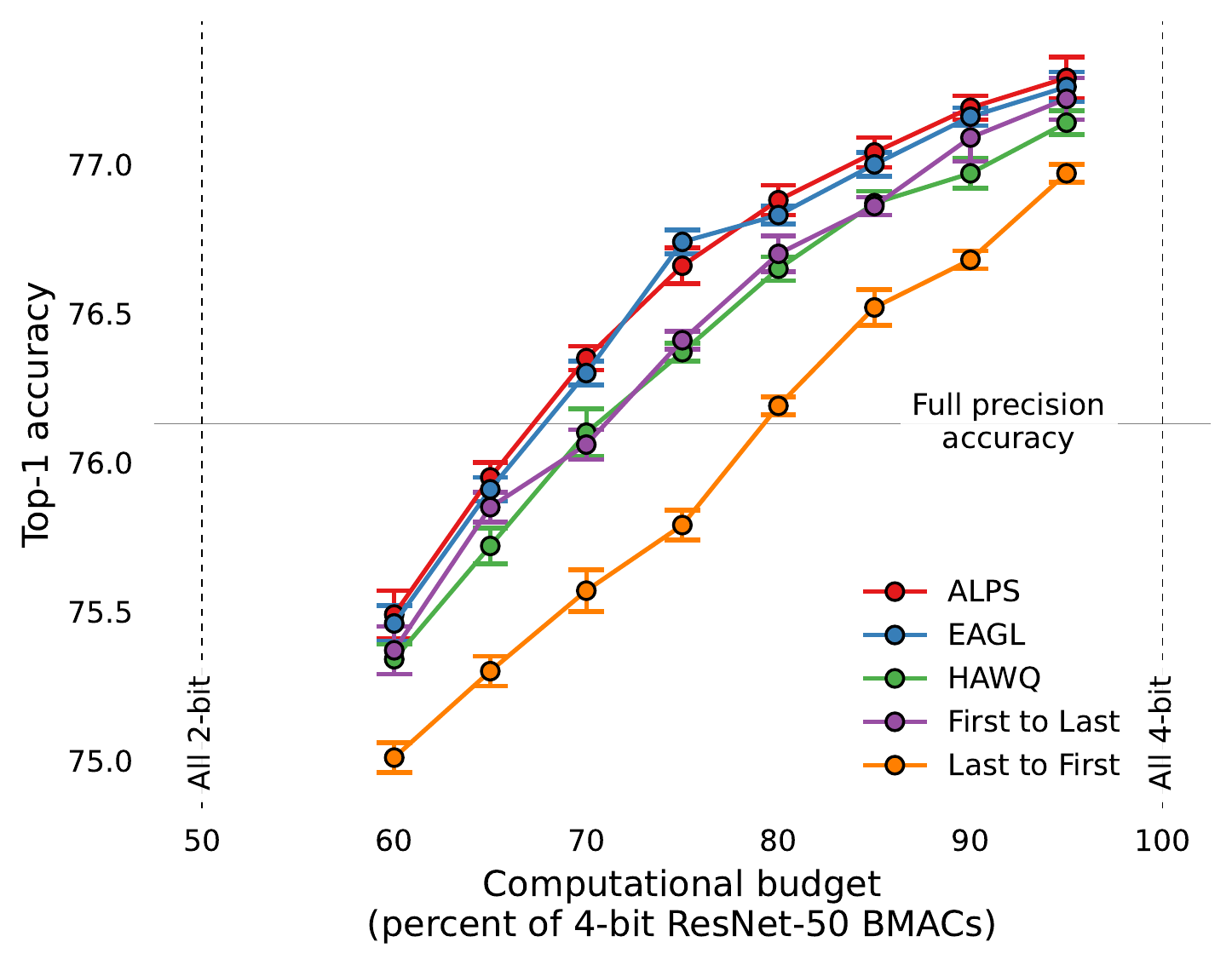}
 \includegraphics[width=\columnwidth]{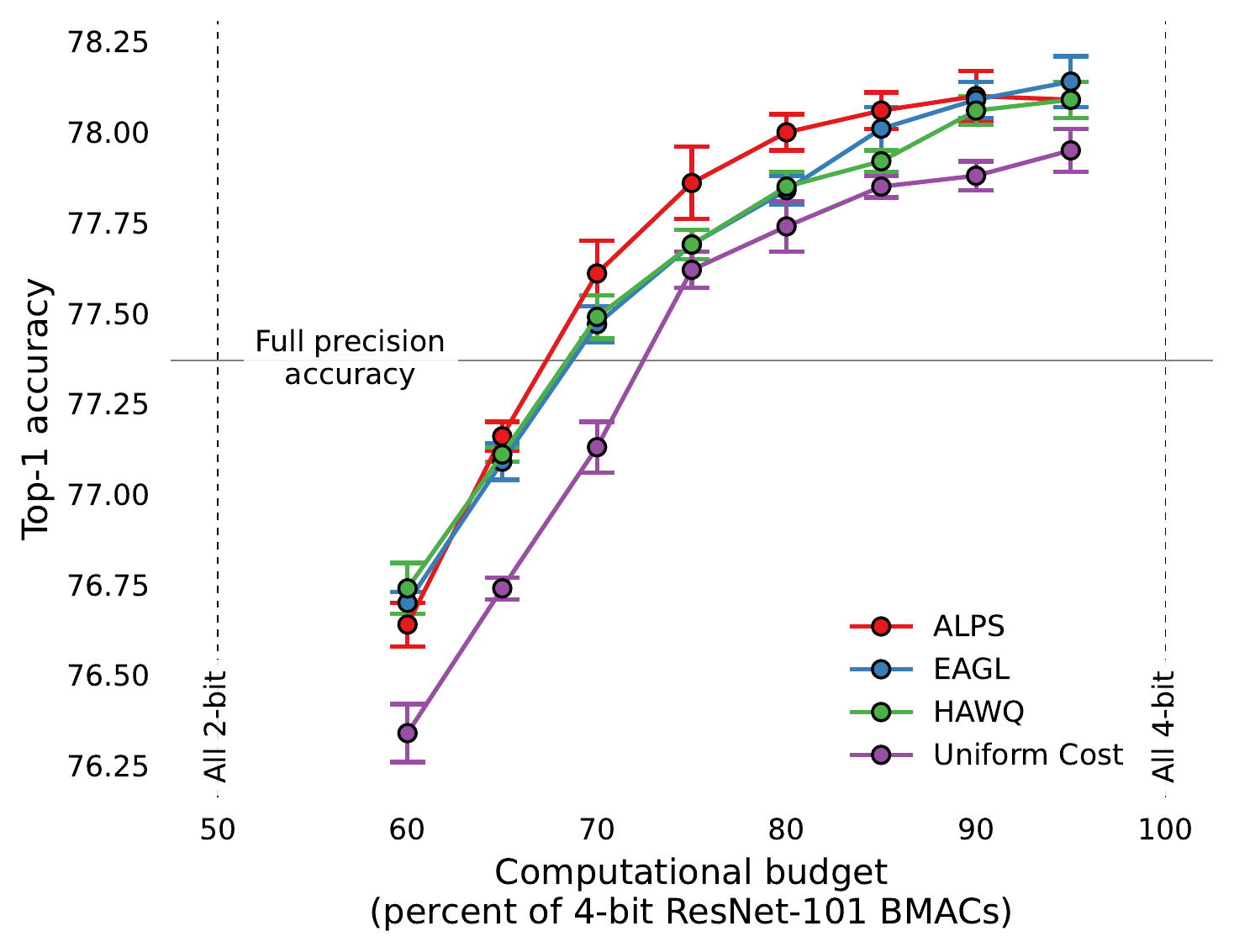}
  \caption{\textbf{ALPS and EAGL perform better than leading mixed precision techniques using ResNet-50 for all computational budgets and meet or exceed the accuracy of ResNet-101 for 7 out of 8 computational budgets on~\cite{deng2009imagenet}.} Mean +/- standard deviation across 5 seeds for each technique at each budget. A network with all configurable layers at 4 bits has a computational budget of 100\% and a network with all configurable layers at 2 bits has a computational budget of 50\% in this plot. The first and last layer are 8-bit and intermediate layers with less than 128 input features are fixed at 4-bit.}
  \label{fig:resnet50and101knapsackfrontier}
\end{center}
\vskip -0.2in
\end{figure*}

Results are shown in Fig.~\ref{fig:resnet50and101knapsackfrontier}. ALPS and EAGL outperform the baselines and HAWQ-v3 at all budgets along the entire throughput-accuracy frontier for ResNet-50, demonstrating very strong performance ($p=0.0079$ for all compute budgets except for EAGL at 60\% where $p=0.0238$, Wilcoxon rank-sum, $N=5$). For ResNet-101, EAGL matches or outperforms HAWQ-v3 at all 8 budgets. ALPS does so at 7 out of 8 budgets. ALPS provides solutions significantly better than HAWQ-v3 and EAGL for 4 of the budgets under consideration. Layer-wise precision selection choices between the five techniques at the 70\% budget for resNet-50 are compared in Appendix \ref{sec:layerwiseapp}. Even with a throughput budget significantly lower than a 4-bit network, with over half of the computation happening at 2-bit instead of 4-bit, EAGL and ALPS match full precision accuracy. For ALPS, one epoch of finetuning for each layer was chosen to minimize computational cost; rerunning the ResNet-50 experiment with two epochs per layer did not improve the result.

The actual computational cost to get the layer-wise estimates for ALPS, EAGL and HAWQ-v3 are listed in Table \ref{bakeoffR50}. EAGL's superior performance given the orders of magnitude saving in terms of computational cost makes it the most suitable candidate for layer selection in practice for getting a fast solution across a range of architectures and budgets. Python code used for computation of the EAGL metric is provided in the Appendix \ref{code}. 
\subsection{Evaluation using PSPNet on semantic segmentation}
\begin{figure}[ht]
\vskip 0.2in
\begin{center}
\centerline{\includegraphics[width=\columnwidth]{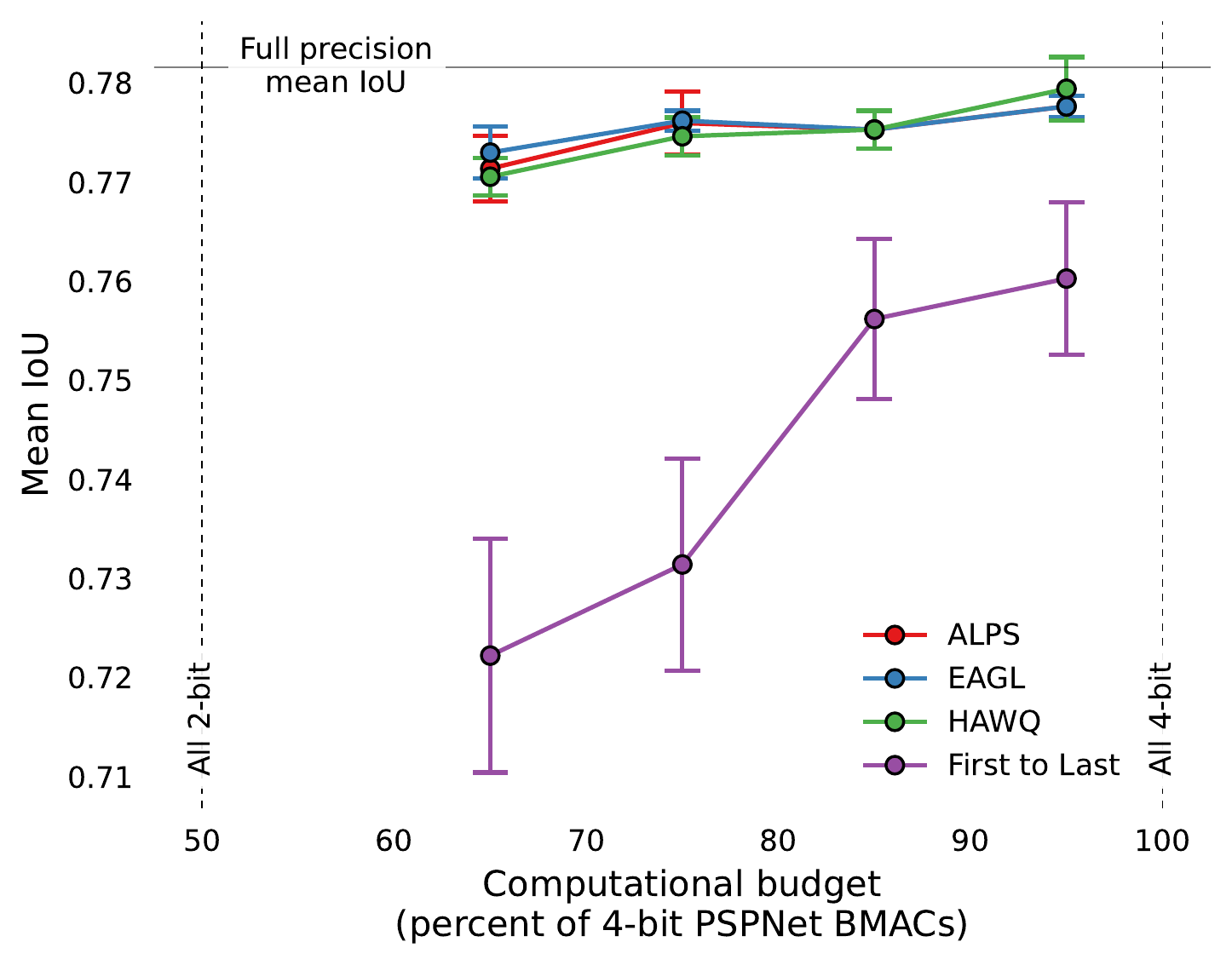}}
  \caption{\textbf{ALPS and EAGL meet or exceed the mean IoU of leading techniques on PSPNet across computational budgets.} Mean +/- standard deviation across 5 seeds at each budget.}
  \label{fig:pspnetknapsackfrontier}
\end{center}
\vskip -0.2in
\vspace{-1.5em}
\end{figure}

Next, ALPS and EAGL are evaluated on the semantic segmentation task~\cite{cordts2016cityscapes} using PSPNet. The methods are compared against HAWQ-v3 and a first to last baseline. Four different target throughput budgets are evaluated, roughly at 95\%, 85\%, 75\%, and 65\% of the computational cost of a 4-bit PSPNet network. For each budget, each technique is fine-tuned for 40,000 iterations using 5 different seeds. Results are shown in Fig. \ref{fig:pspnetknapsackfrontier}. 
%\begin{table}[h]
%\caption{\textbf{Metric computational cost comparison for PSPNet. EAGL and ALPS are much faster %to compute than other techniques.} The cost compared here includes only the computational %resources required for the estimation of the layer-wise metric for each method, and not the %computation cost for fine-tuning the mixed precision network, nor the cost for training the %initial 4-bit checkpoint, as these are the same for all techniques.}
%\label{bakeoffPSPNet} 
%\vskip 0.15in
%\begin{center}
%\begin{small}
%\begin{sc}
%\begin{tabular}{lcccr}
%\toprule
%Method  & Metric Compute Cost \\
%\midrule
%\textbf{EAGL (Ours)}    & \textbf{<1 CPU minute} \\ \hline
%ALPS (Ours)     & 67 GPU hours\\ \hline
%HAWQ-v3 & ~1032 GPU hours \\
%\bottomrule
%\end{tabular}
%\end{sc}
%\end{small}
%\end{center}
%\vskip -0.1in
%\end{table}

The task performance, i.e., mean Intersection over Union (IoU) for networks produced by ALPS and EAGL is not significantly different from HAWQ-v3 ($p>0.1$ for all compute budgets, Wilcoxon rank-sum test, $N=5$), while all methods outperformed the first-to-last baseline ($p=0.0079$ for all budgets, Wilcoxon rank-sum, $N=5$). The computational cost to get the layer-wise estimates for ALPS, EAGL and HAWQ-v3 are listed in Table \ref{bakeoffR50}. EAGL and ALPS are able to find good solutions to the mixed precision layer selection problem for this network, while being orders of magnitude faster to compute and easier to implement in practice. The results provide further evidence that both methods are scalable, task agnostic approaches that generalize to other networks and datasets as well. 

\subsection{Evaluation using BERT-base model on the SQuAD1.1 dataset}
To further evaluate generalizability to other application domains, ALPS and EAGL are evaluated on the SQuAD1.1 question answering benchmark, using the BERT-base transformer model. The methods are compared against two baseline layer selection techniques. The first baseline ranks layers from first to last based on their topological order and drops the precision of the first n layers greedily until the resulting network just meets the budget. The second baseline follows the same approach except that layers are ranked from last to first based on their topological order.

Four distinct target budgets are sampled for evaluation, roughly at 90\%, 80\%, 70\%, and 60\% of the computational cost of a 4-bit network. The input to the softmax layer is fixed at 8-bit. For each budget, for each technique, networks are fined-tuned for 60 epochs using 3 seeds. 

\begin{figure}[ht]
\vskip 0.2in
\begin{center}
\centerline{\includegraphics[width=\columnwidth]{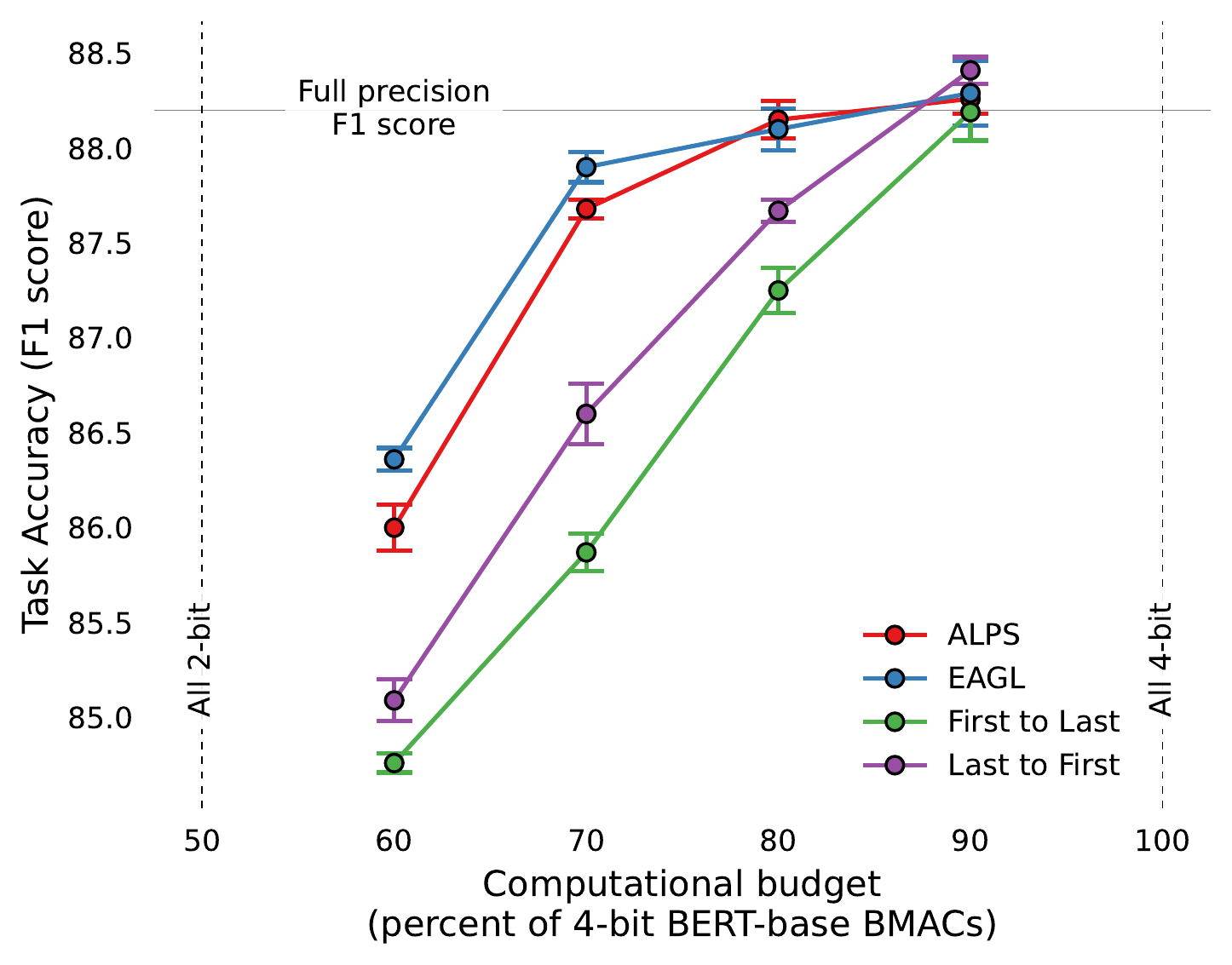}}
  \caption{\textbf{ALPS and EAGL find more accurate mixed precision networks for SQuAD v1.1 across all computational budgets.} Mean +/- standard deviation across 3 seeds at each budget.}
  \label{fig:bertbasefigure}
\end{center}
\vskip -0.2in
\vspace{-1.5em}
\end{figure}

Models for all approaches are fine-tuned using LSQ, starting from a 4-bit checkpoint created using FAQ. Results are shown in Fig. \ref{fig:bertbasefigure}. EAGL\footnote{As EAGL relies on a trained checkpoint for its layer-wise estimates, and since we use LSQ for finetuning the mixed precision networks, the entropy estimates for EAGL were computed on a 4-bit checkpoint created using LSQ, as this was found to give better results than using the 4-bit FAQ checkpoint for this purpose.} and ALPS find networks which have better or equal performance across the entire accuracy-throughput frontier on this Natural Language Processing task, providing evidence for generalizability across different architectures and domains.

\section{Conclusion and Discussion}
Two new metrics for optimizing precision settings in a mixed precision network are introduced. Both outperform leading methods from the literature while requiring fewer computational resources to calculate and generalize across different tasks and architectures. Further evidence for the quality of these metrics is provided in Appendix \ref{Sec:solutionquality}. 

The ALPS metric is intuitive, easy to implement and yet outperforms leading techniques from literature that ignore fine-tuning and use complex proxies which don't scale well in practice, both in terms of task performance of the proposed solution and time needed to compute the metric. The entropy-based EAGL metric is elegant, and applicable even when only the model parameters are known but the training data is inaccessible, making it task agnostic and generalizable to other domains like unsupervised learning. Since it only requires access to a trained checkpoint, EAGL is remarkably fast, and is shown to select more accurate networks than several leading state-of-the-art methods on several datasets and networks in a fair comparison.

Given its strong performance, extremely low computational cost and ease of implementation, EAGL provides effective, practically usable solutions for network compression and faster inference. This makes EAGL the first choice to get good solutions extremely fast. The performance can sometimes be further improved by using ALPS. Although this paper considers networks consisting only of 2- and 4-bit layers, both methods can be used with more than two precision choices by changing the optimizer (e.g.~\cite{chen2021towards}). Given the strong results across the entire throughput-accuracy frontier, as the number of hardware platforms supporting mixed precision grows, these simple, efficient, and effective approaches should be of great utility to practitioners, directly allowing the deployment of lower power, higher throughput solutions each time they are used. 
\section{Acknowledgement}
This material is based upon work supported by the United States Air Force under Contract No. FA8750-19-C-1518.
{\small
\bibliographystyle{icml2023}
\bibliography{main}
}

%%%%%%%%%%%%%%%%%%%%%%%%%%%%%%%%%%%%%%%%%%%%%%%%%%%%%%%%%%%%%%%%%%%%%%%%%%%%%%%
%%%%%%%%%%%%%%%%%%%%%%%%%%%%%%%%%%%%%%%%%%%%%%%%%%%%%%%%%%%%%%%%%%%%%%%%%%%%%%%
% APPENDIX
%%%%%%%%%%%%%%%%%%%%%%%%%%%%%%%%%%%%%%%%%%%%%%%%%%%%%%%%%%%%%%%%%%%%%%%%%%%%%%%
%%%%%%%%%%%%%%%%%%%%%%%%%%%%%%%%%%%%%%%%%%%%%%%%%%%%%%%%%%%%%%%%%%%%%%%%%%%%%%%
\newpage
\onecolumn
\appendix
\section{{Demonstration of additivity of layer-wise accuracy estimates}} \label{Sec:additive}

\begin{figure*}[t]
\centering
\includegraphics[width=0.5\textwidth, ,height=0.4\textwidth]{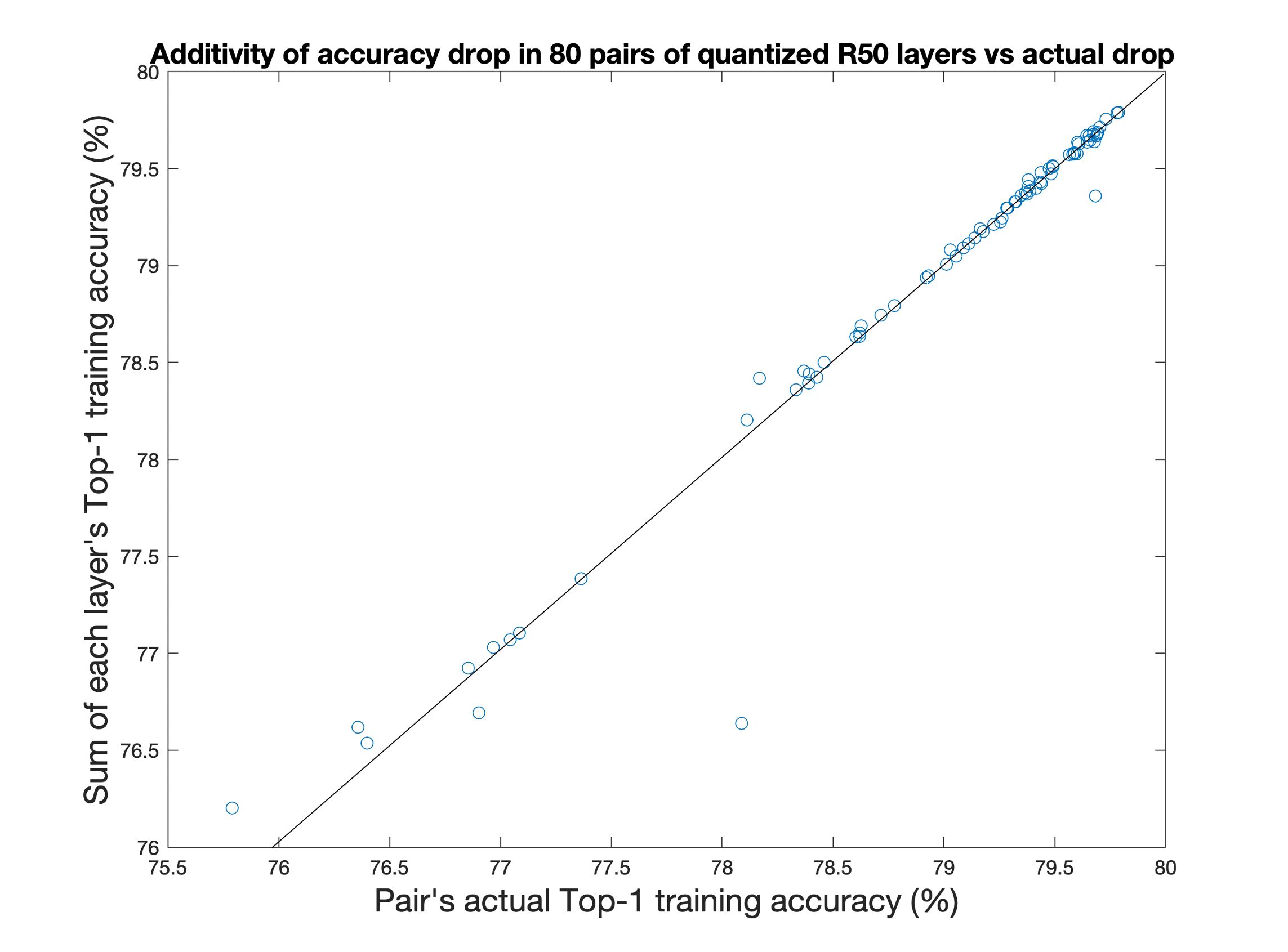}
  \caption{\textbf{Accuracy drop in 80 random pairs of ResNet-50 layers is additive, justifying the assumptions of the optimization methodology (See text for details).}}
  \label{fig:additivity}
\end{figure*}

Two experiments were conducted to test the assumption that layerwise accuracy estimates can be combined linearly to determine overall network accuracy.  This assumption is implicit in the use of the knapsack optimization algorithm.

The first experiment directly tests whether the drop in accuracy when quantizing two layers at a time, $L_1$ and $L_2$, is the same as the sum of accuracy decline when each is quantized alone.  Specifically start with a 4-bit ResNet-50 network fine-tuned with Quantization Aware Training as described in the methods. Then, for each layer, measure $D(L)$, the Top-1 training set accuracy drop when that layer's precision is dropped from 4-bit to 2-bit, with no further fine-tuning.
For 80 random pairs of layers, $<L_1,L_2>$,
1) predict the accuracy drop as $D(L_1) + D(L_2)$, and 2)
measure the actual drop in training set accuracy with both layers reduced to 2 bits, with no further fine-tuning. Figure \ref{fig:additivity} plots the predicted versus actual accuracy drop.  The two measures are strongly correlated (R=0.98), indicating that the assumption of linearity required by the optimization methodology is justified.

The second experiment is a more stringent test of linearity; instead of testing pairs-wise linearity, it tests the linearity of arbitrary combinations of layers by testing whether an accurate linear regression model can be constructed to predict the network accuracy given only the precision choices for all layers.

Specifically, to test if a linear regression model can predict the accuracy of mixed precision models given the layer-wise precision settings:
1) Train 135 stratified random mixed-precision ResNet-50 networks for 30 epochs as described in the methods section, i.e. 3 networks each for 2, 3, …, 46 randomly chosen layers at 2-bit, in an otherwise 4-bit network.
2) For each training run record a) the precision of each layer as 48-element binary vector, where 0 and 1 indicate 2 and 4 bits, respectively, and b) the final accuracy on the validation set.
3) Divide the 135 samples into training/hold-out set randomly (90\%/10\%).
4) Perform linear regression on the training set.
5) Use the model to predict the network accuracy on the training samples and hold-out set to compare with the actual scores.

\begin{figure*}[t]
\centering
\includegraphics[width=0.9\textwidth, ,height=0.4\textwidth]{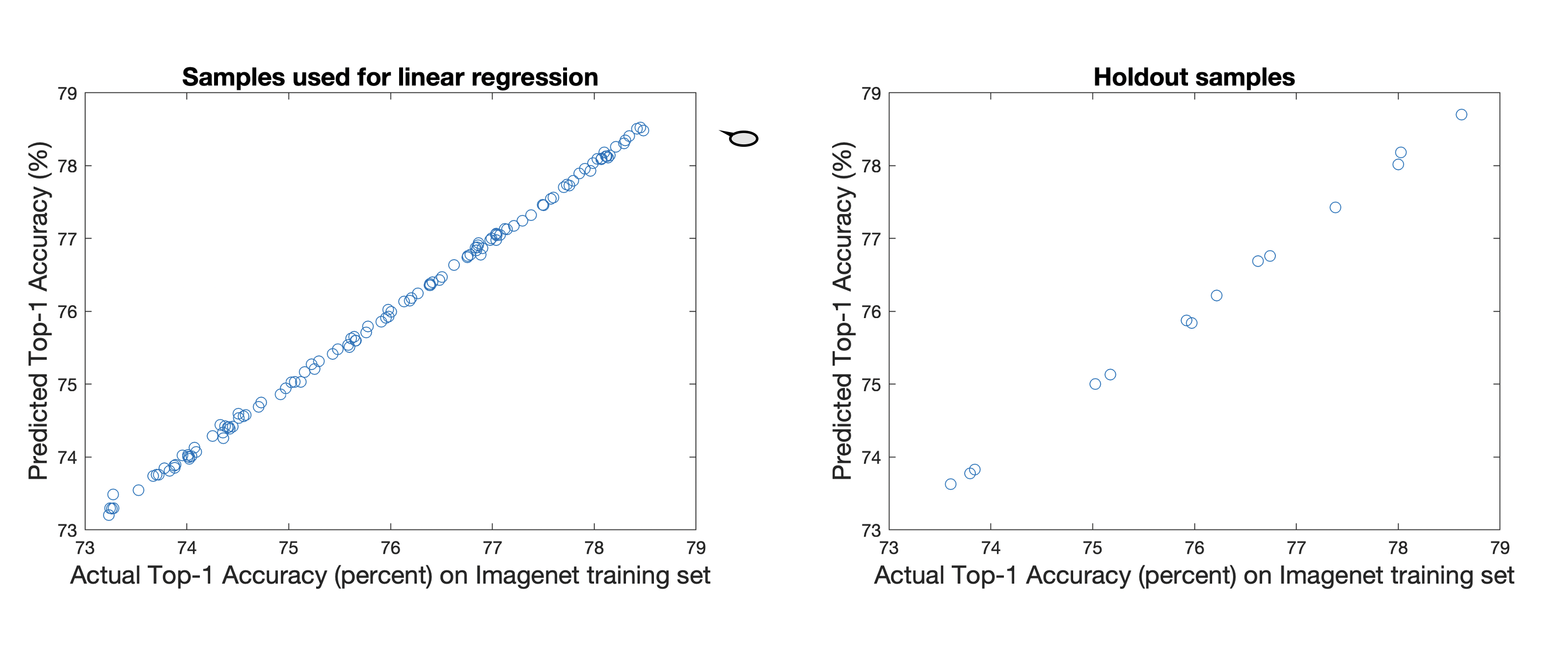}
  \caption{\textbf{A linear regression model accurately predicts the overall network accuracy given the individual layer precision settings, further justifying the linearity assumption of the optimization methodology (See text for details).}}
  \label{fig:linearregression}
\end{figure*}

Figure \ref{fig:linearregression} shows the actual Top-1 accuracy on the benchmark versus the linear regression model's prediction for the samples used to create the model (left), and the hold-out set (right).  The figure shows that the overall network accuracy can be modeled very well as a linear combination of the individual layer accuracy contributions (R=0.9996 on the model training data; R=0.9994 on the hold-out data).  This provides further evidence for linearity required by the knapsack solver used herein. 

\section{On the quality of the EAGL and ALPS solutions} \label{Sec:solutionquality}
It would be ideal to be able to compare the EAGL and ALPS solutions against a ground-truth measure.  Although exhaustive search to provide such is infeasible, an additional experiment was conducted using ResNet-50 in an attempt to measure how well the EAGL and ALPS solutions compare with the strongest, although impractical, layer selection metric that we could construct.

Given the accuracy of the regression model in Section~\ref{Sec:additive} in determining the overall accuracy of the network given the layer precisions, the linear coefficient for layer $l$ in the model was used as $G_l$, providing an alternative layer selection metric. The frontier was then found using the same knapsack optimization procedure as for EAGL and ALPS described above. Each method was trained for 30 epochs at each budget (N=6 for each datapoint). It can be argued that this provides very good solutions since 1) As was shown in Figure \ref{fig:linearregression} and described in the previous section, the regression model coefficients accurately reflect the layer-wise contributions to total network accuracy (hold-out set residual error standard deviation=0.068\%), and 2) the knapsack optimizer will provide an $\epsilon$-optimal solution to the problem subject to the quality of the layerwise accuracy estimates.

\begin{figure*}[t]
\centering
\includegraphics[width=0.9\textwidth, ,height=0.5\textwidth]{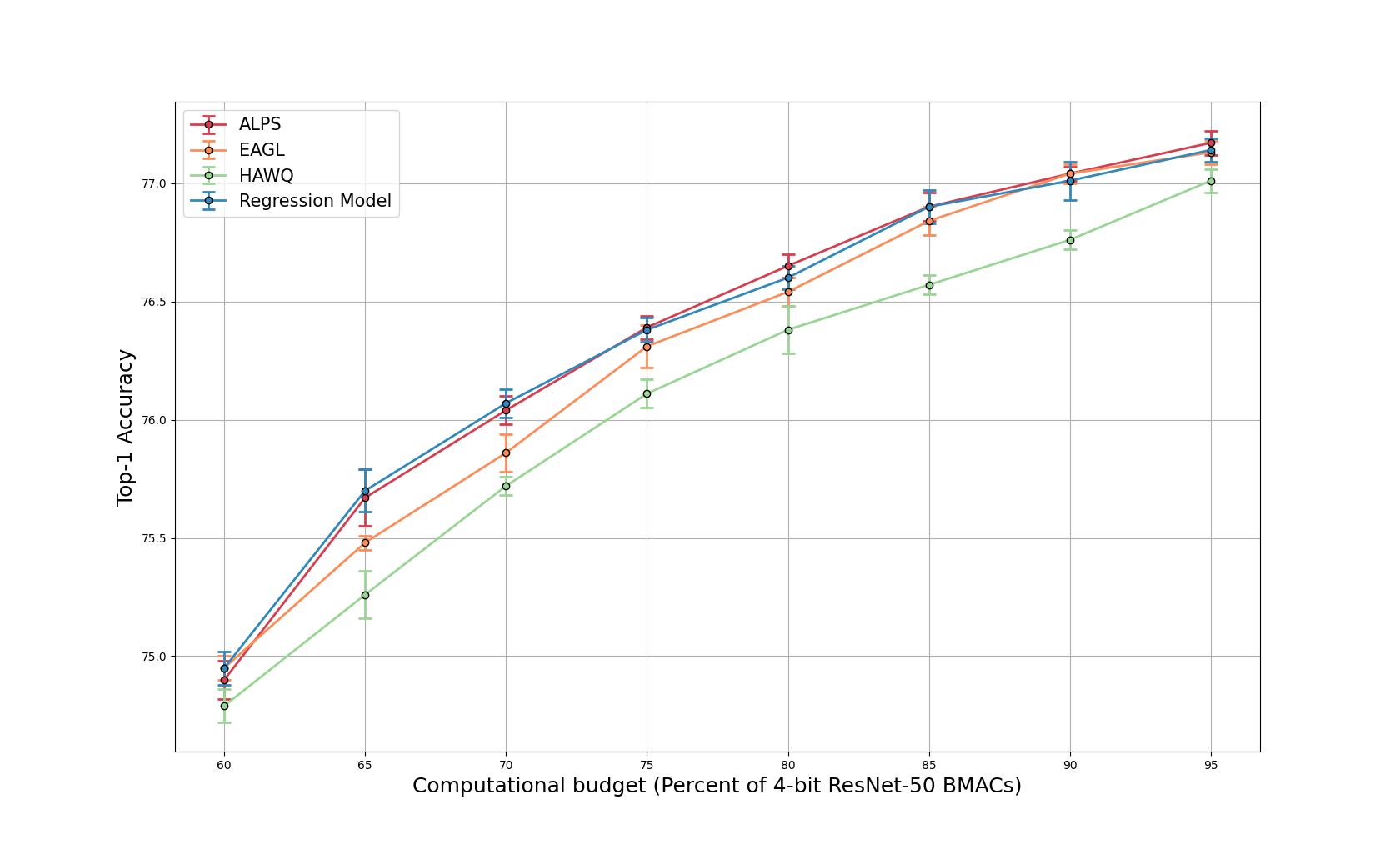}
  \caption{\textbf{EAGL and ALPS frontiers are very close to a frontier generated using the coefficients from an accurate regression model on ResNet-50, a computationally expensive, but arguably stronger accuracy-aware method (see text for details).}}
  \label{fig:regression_model_frontier}
\end{figure*}

Figure \ref{fig:regression_model_frontier} shows the results of this experiment, where the mean and standard-deviation of the Top-1 accuracy of mixed-precision networks are shown for each of 8 computational budgets.  Note that ALPS and EAGL are very close to the frontier provided by the accurate regression model, supporting the quality of the solutions provided by the proposed methods, which are computationally tractable \footnote{It took approximately 1,080 A100 GPU hours to create the regression model for ResNet-50, making it impractical as a layer selection method in practice.}. Perhaps there is little room for improvement beyond that of ALPS and EAGL.

\section{Re-implementation of HAWQ-v3} \label{Sec:hawqdetails}
To compare EAGL and ALPS results with those of HAWQ-v3 in a 2- and 4-bit mixed precision network, it was necessary to use the author's implementation for comparison, since 2/4 bit results were not reported for ResNet50 with the constraint that both weights and inputs to a convolutional layer had to have the same precision. Furthermore, no results were reported for PSPNet.  

For a commensurate comparison, HAWQ-v3 was evaluated using the same initial checkpoint and network used in our ALPS and EAGL experiments. Additionally, the 0-1 Knapsack optimization algorithm was used to make precision choices given the layer-wise metric used in ~\cite{dong2019hawq}.  Specifically, their Pyhessian implementation ~\cite{yao2020pyhessian} was used to calculate the average Hessian trace of each layer and estimate the accuracy gain for keeping layers at 4-bit as 
\[ \overline{Trace(H)}||Q_4(W)-Q_2(W)||_2^2\]
where \(\overline{Trace(H)} \) is the average of the diagonal of the Hessian, \(Q_b(W)\) is the tensor, \(W\), quantized using \(b\)-bit, as in section 3.  When lowering the precision of each layer's weights from 4 to 2 bits, the author's step-size initialization method of using the range of the weights divided by \(2^{precision-1}\) was followed; however to keep the uniform quantization intervals symmetric about 0, we adjust the range to be between \(\pm~max({|min(W)|,|max(W)|)}\) 

\section{Layer precision selection comparison on ResNet-50} \label{sec:layerwiseapp}
\begin{figure*}[ht]
\centering
\includegraphics[width=1\textwidth, ,height=0.33\textwidth]{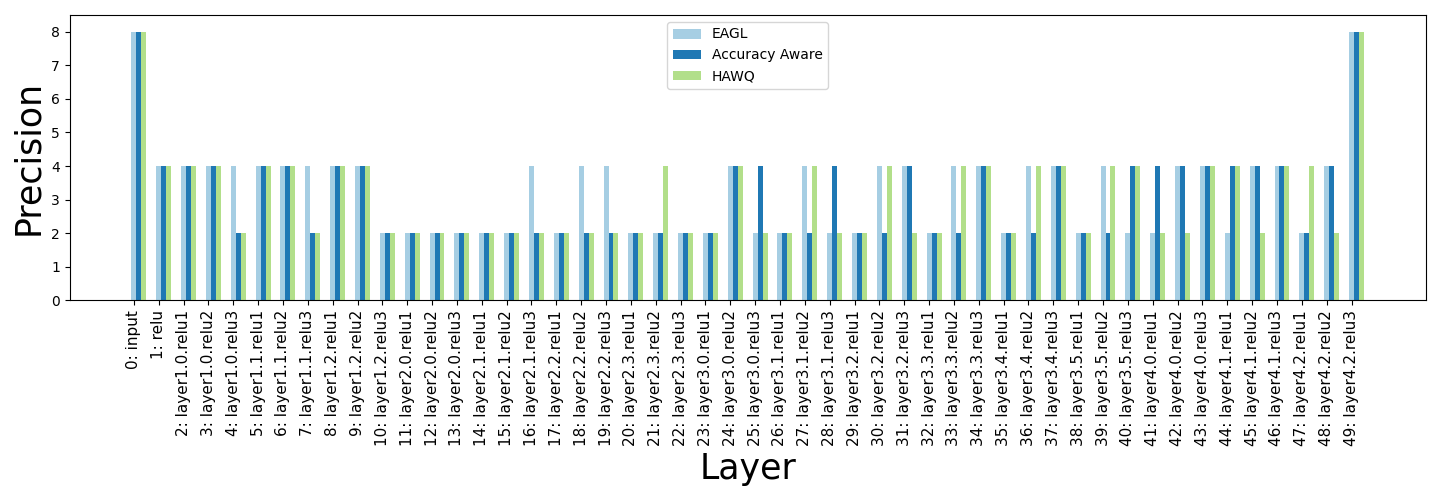}
  \caption{\textbf{Layerwise precision selection comparison between techniques.} Some techniques choose fewer total number of layers to quantize to 2 bits than others. The total number of layers chosen does not necessarily predict the final accuracy of the technique. Downsample layers are omitted in this plot. The precision of each downsample layer is the same as the convolutional layer which connect to the same ReLU as the corresponding downsample layer.}
  \label{fig:r50layerwise}
\end{figure*}
Layer-wise precision selections are compared between techniques in Figure \ref{fig:r50layerwise}. EAGL drops fewer layers to 2-bit for the same computational budget than HAWQ-v3 and ALPS, which drop the same number of layers. Both ALPS and EAGL perform much better at this budget than HAWQ-v3, demonstrating that the final accuracy is not merely a function of keeping the fewest number of layers at 2-bit.

\newpage
\section{Python code to calculate EAGL metric}\label{code}

The PyTorch code snippet below calculates the EAGL metric on a trained low precision checkpoint. The precision and scale information for the weight tensor is assumed to be stored in the checkpoint.
%\begin{lstlisting}
\begin{python}
def EntropyBits(p):
    ent = -sum(p[i] * log2(p[i]) 
              for i in range(len(p)))
    return ent

def quantized_tensor(tensor, scale, precision):
    qt = tensor/scale
    minval = -2**(precision-1)
    maxval = 2**(precision-1)-1
    qt = qt.clamp(minval,maxval)
    qt = qt.round()
    return qt
    
for key in checkpoint['state_dict'].keys():
    if 'weight' in key:
        inten = checkpoint['state_dict'][key]
        scale = checkpoint['state_dict'] 
            [key+'_scale']
        precision = checkpoint['state_dict'] 
            [key+'_precision']
        qt = quantized_tensor(inten, scale, 
            precision)
        minlength = 2**precision 
        newval=(qt.contiguous().view(-1)+
            (-layer_min)).int() 
        px=torch.bincount(newval,minlength=
            minlength).float()/newval.numel()
        layer_entropy = EntropyBits(px+1e-10)

\end{python}
\end{document}